\pdfoutput=1

\documentclass[11pt]{article}

\usepackage{EMNLP2022}

\usepackage{times}
\usepackage{latexsym}

\usepackage[T1]{fontenc}

\usepackage[utf8]{inputenc}

\usepackage{microtype}

\usepackage{inconsolata}

\usepackage{longtable}
\usepackage{verbatim}
\usepackage{lipsum}
\usepackage{hyperref}
\usepackage{graphicx}
\usepackage{multirow}
\usepackage{array}
\newcommand{\STAB}[1]{\begin{tabular}{@{}c@{}}#1\end{tabular}}

\usepackage{soul}
\usepackage{color}
\newcommand{\hlc}[2][yellow]{ {\sethlcolor{#1} \hl{#2}} }

\title{Multitask Instruction-based Prompting for Fallacy Recognition}

\author{Tariq Alhindi$^1$ \hspace{0.5cm}
        Tuhin Chakrabarty$^1$ \hspace{0.5cm}
        Elena Musi$^3$ \hspace{0.5cm}
        Smaranda Muresan$^{1,2}$\\
  $^1$Department of Computer Science, Columbia University\\
  $^2$Data Science Institute, Columbia University\\
  $^3$Department of Communication and Media, University of Liverpool\\
  {\normalsize \tt \{tariq, tuhin.chakr, smara\}@cs.columbia.edu} \\
  {\normalsize \tt elena.musi@liverpool.ac.uk}}

\begin{document}
\maketitle
\begin{abstract}
Fallacies are used as seemingly valid arguments to support a position and persuade the audience about its validity. Recognizing fallacies is an intrinsically difficult task both for humans and machines. Moreover, a big challenge for computational models lies in the fact that fallacies are formulated differently across the datasets with differences in the input format (e.g., question-answer pair, sentence with fallacy fragment), genre (e.g., social media, dialogue, news), as well as types and number of fallacies (from 5 to 18 types per dataset). To move towards solving the fallacy recognition task, we approach these differences across datasets as multiple tasks and show how instruction-based prompting in a multitask setup based on the T5 model improves the results against approaches built for a specific dataset such as T5, BERT or GPT-3. We show the ability of this multitask prompting approach to recognize 28 unique fallacies across domains and genres and study the effect of model size and prompt choice by analyzing the per-class (i.e., fallacy type) results. Finally, we analyze the effect of annotation quality on model performance, and the feasibility of complementing this approach with external knowledge.

\end{abstract}

\begin{table}[]
    \small
    \centering
    \begin{tabular}{|l|}
    \hline
        
        \rule{0pt}{\normalbaselineskip} \hspace{-.2cm}
        \textbf{Question-Answering dialog moves in \textsc{Argotario}:}\\
         Has anyone been on the moon?\\
         The moon is so far away, we should focus on our society. \\
         
         \rule{0pt}{\normalbaselineskip} \hspace{-.2cm} 
         \textbf{Fallacy:} \hlc[Goldenrod]{Red Herring} \\
    \hline
        
        \rule{0pt}{\normalbaselineskip} \hspace{-.2cm}
        \textbf{Propaganda techniques in news:}\\
         The ability to build an untraceable, unregistered gun \\
         is definitely a \underline{game changer.} \\
         
         \rule{0pt}{\normalbaselineskip} \hspace{-.2cm}
        \textbf{Fallacy:} \hlc[Goldenrod]{Loaded Language} \\
    \hline
        
        \rule{0pt}{\normalbaselineskip} \hspace{-.2cm}
        \textbf{Educational website on fallacies:}\\
          She is the best because she is better than anyone else\\
         
         \rule{0pt}{\normalbaselineskip} \hspace{-.2cm}
        \textbf{Fallacy:} \hlc[Goldenrod]{Circular Reasoning} \\
    \hline
            
        \rule{0pt}{\normalbaselineskip} \hspace{-.2cm}
        \textbf{Fact-checked news:}\\
          Says Joe Biden has said 150 million Americans died \\
          from guns and another 120 million from COVID-19.\\
         
         \rule{0pt}{\normalbaselineskip} \hspace{-.2cm}
        \textbf{Fallacy:} \hlc[Goldenrod]{Cherry Picking} \\
    \hline
    
    \end{tabular}
    \caption{Examples of fallacies from multiple datasets}
    \label{tab:example}
\end{table}

\section{Introduction}
\label{sec:intro}
A fallacious argument is one that seems valid but it is not \cite{hamblin2022fallacies}. 
Theoretical work in argumentation has introduced various typologies of fallacies. For example, 
\newcite{van2002argumentation} consider 
fallacies that occur when an argument violates 
the ten rules of a critical discussion, while  \newcite{tindale2007fallacies} categorizes fallacies into 4 categories: 
structural fallacies, related to the number and structure of arguments; fallacies from diversion, drawing from the (un)intentional diversion of the attention from the issue at hand; logical fallacies, related to the argument scheme at play and language fallacies, 
related to vagueness or ambiguity. 
Fallacious reasoning can bring misbehaviour and be used for manipulation purposes. Thus, having a system that can recognize fallacy types across domains and genres is crucial for applications that teach humans how to identify fallacies and avoid using them in their arguments. 

Work in computational models for fallacy recognition is still in its infancy, with a limited set of relatively small datasets such as \textsc{Argotario} \cite{habernal-etal-2017-argotario}, which consists of question and answer dialog moves; name-calling in social media debates \cite{habernal-etal-2018-name}, fallacies as propaganda techniques in news \cite{da-san-martino-etal-2019-fine}; logical fallacies from educational websites  \cite{jin2022logical}, and fallacies used for misinformation in social media and news around Covid-19 \cite{musi2022developing}. Table \ref{tab:example}, shows some examples of fallacies from these datasets. 

Previous work on fallacy recognition has tackled just one dataset at a time. For example, work on detecting propaganda techniques use fine-tuning of different pre-trained transformers with embedding-based or handcrafted features \cite{da-san-martino-etal-2020-semeval,jurkiewicz-etal-2020-applicaai} as well as LSTMs and transformers for sequence tagging of propaganda fragments \cite{da-san-martino-etal-2019-findings,yoosuf-yang-2019-fine,alhindi-etal-2019-fine,chernyavskiy-etal-2020-aschern}, while \newcite{jin2022logical} propose a structure-aware classifier to detect logical fallacies. 

Fallacy recognition is a challenging task for three main reasons: i) the number of classification labels (fallacy types) and class imbalance in existing datasets is often very high; ii) existing datasets cover varying genres and are typically very small in size due to annotation challenges; and iii) models trained on individual data sets often show poor out of distribution generalization. 

A recent line of work \cite{wei2021finetuned,sanh2022multitask} relies on the intuition that most natural language processing tasks can be described via natural language instructions and models trained on these instructions in a multitask framework show strong zero-shot performance on new tasks.  
Based on this success, we propose a \textit{unified model based on multitask instruction-based prompting} using T5 \cite{raffel2020exploring} to solve the above challenges for fallacy recognition (Section \ref{sec:method}). This approach allows us to unify all the existing datasets and a newly introduced dataset (Section \ref{sec:data}) by converting 28 fallacy types across 5 different datasets into natural language instructions. In particular, we address the following research questions: i) Can we have a unified framework for fallacy recognition across domains, genres, and annotation schemes? ii) Are fallacy types expressed differently across datasets? iii) What are the effects of model size and prompt choice on the per-class performance for the fallacy recognition task?

Experimental evidence shows that our multitask fine-tuned models outperform task specific models trained on a single dataset by an average margin of 16\%  as well as beat strong 
few-shot and zero-shot baselines by average margins of 25\% and 40\%, respectively in macro F1 scores across five datasets (Section \ref{sec:res}). To further deepen our understanding towards the task of fallacy recognition we analyze the performance of our models for each fallacy type across datasets, model size and prompt choice (Section \ref{ssec:fal_type}). We further analyze the effect of annotation quality on the model performance, and the feasibility of complementing this approach with external knowledge (Section \ref{ssec:error}). We make all datasets, code and models publicly available.\footnote{\url{https://github.com/Tariq60/fallacy-detection}}

\section{Data}
\label{sec:data}
We experiment with five datasets (4 existing and a new dataset) 
that cover 28 unique fallacy types  
in multiple domains (e.g., covid-19, climate change, politics) and genres (e.g. news articles, QA turns in dialog, social media). Table \ref{tab:data} shows a summary of all datasets (65\% training, 15\% development, 20\% test) and Table \ref{tab:counts} in Appendix \ref{app:fal} shows detailed counts of fallacy types in each split for all datasets.

\begin{table}[]
    \scalebox{.8}{
    \centering
    \begin{tabular}{|l|l|l|l|l|}
         \hline
         Data &Ex &F &Genre &Domain\\
         \hline
        \textsc{Argotario}  &880 &5 &Dialogue &General\\
        \textsc{Propaganda} &5.1k &15\textsuperscript{$\dagger$} &News &Politics\\
        \textsc{Logic}      &4.5k &13 &Diverse &Education\\
        \textsc{Covid-19}   &621 &9\textsuperscript{$\ddagger$} &SocMed/News 
        &Covid-19\\
        \textsc{Climate}    &477 &9\textsuperscript{$\ddagger$} &News
        &Climate\\
         \hline
    \end{tabular}}
    \caption{Summary of five fallacy datasets. \textbf{Ex}: Total number of examples. \textbf{F}: Final number of fallacy types after unifing all datasets. \textsuperscript{$\dagger$}Original scheme has 18 propaganda techniques.  \textsuperscript{$\ddagger$}Original scheme has 10 fallacy types.
    }
    \label{tab:data}
\end{table}

\paragraph{Existing Fallacy Datasets} 
We include four existing fallacy datasets in our experiments.

The first dataset is \textsc{Argotario}, introduced by \citet{habernal-etal-2017-argotario}, a dataset for fallacy detection where given a QA pair the task is to detect the fallacy in answers. 
Their scheme include five fallacy types: \textit{Ad Hominem, Appeal to Emotion, Red Herring, Hasty Generalization, Irrelevant Authority}. 

The second dataset (\textsc{Propaganda}) contains 18 propaganda techniques in news articles annotated at the fragment and sentence levels 
\cite{da-san-martino-etal-2019-fine}. We focus on 15  that are fallacies and frequent enough in the data: \textit{Loaded Language, Name Calling or Labeling, Exaggeration or Minimization, Doubt, Appeal to Fear/Prejudice, Flag-Waving, Causal Oversimplification, Slogans, Appeal to Authority, Black-and-White Fallacy, Thought-Terminating Cliche,  Whataboutism, Reductio ad Hitlerum, Red Herring, and Strawman}. 

The third dataset (\textsc{Logic}) is recently released by \citet{jin2022logical} and contains 13 logical fallacies (\textit{Faulty Generalization, False Causality, Circular Claim, Ad Populum, Ad Hominem, Deductive Fallacy, Appeal to Emotion, False Dilemma, Equivocation, Fallacy of Extension, Fallacy of Relevance, Fallacy of Credibility, Intentional Fallacy}) from educational websites on fallacy such as Quizziz and study.com. They contain diverse types of text such as dialogue and short statements (e.g., the \textit{Circular Reasoning} example shown in Table \ref{tab:example}). The authors also introduce another challenge dataset: \textsc{ClimateLogic} that follows the same fallacy scheme. However, it contains text segments that are too long (e.g. multiple paragraphs) with no annotations of smaller fallacious fragments like the Propaganda dataset. Therefore, \textsc{ClimateLogic} is beyond the scope of this study. 

The final existing fallacy dataset (\textsc{Covid-19}) is about fact-checked content around Covid-19 \cite{musi2022developing}. The authors identify 10 fallacies (\textit{Evading the Burden of Proof, Cherry Picking, Strawman, Red Herring, False Authority, Hasty Generalization, Post Hoc, False Cause, False Analogy, Vagueness}) 
through analysis of fact-checked social media posts and news by considering fallacies as indicators of misinformation. 

More detailed description of all datasets 
is shown in Appendix \ref{app:fal}.

\begin{figure*}[t!]
    \centering
    \includegraphics[width=\linewidth]{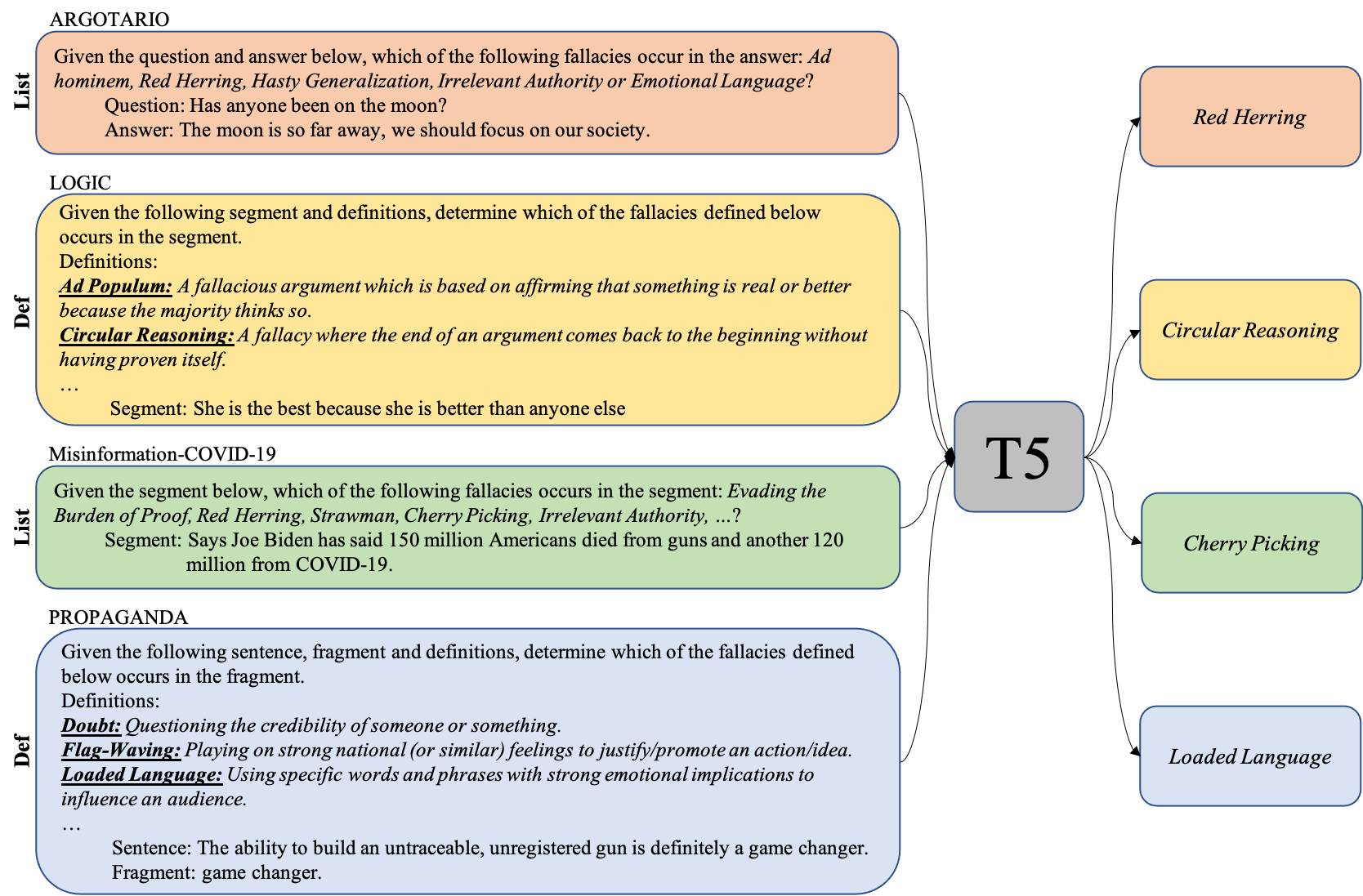}
    \caption{Model and Prompts. \textbf{Def}: fallacy definitions in the prompt.  \textbf{List}: fallacy names listed in the prompt.}
    \label{fig:model}
\end{figure*}

\paragraph{New Fallacy Dataset} 
Drawing from the annotation scheme developed by \citet{musi2022developing}, we annotate 778 segments (477 fallacious) from 92 climate change articles fact-checked by climate scientists at \url{climatefeedback.org}. Each fact-checked article is accompanied by an ``annotations'' section where segments from the original articles are directly followed by the reviewers’ comments.  
Two annotators look at both segments and comments to annotate fallacy types. 
They had a 0.47 Cohen’s $\kappa$ \cite{cohen1960coefficient}, which corresponds to moderate agreement. The gold labels were then done by an expert annotator (in argumentation and fallacy theory) that went over both cases of agreement and disagreement to decide the final label. We denote this dataset as \textsc{Climate} where it differs from \textsc{ClimateLogic} \cite{jin2022logical} in three ways: i) it is built using a fallacy scheme specifically developed for misinformation; ii) the fallacious segments are identified by domain experts at \url{climatefeedback.org} and contain comments which explain fallacious aspects; iii) the segments are mostly 1-3 sentences long.


\paragraph{Final Labels.} 
We unify the labels of similar fallacies (e.g., \textit{False Cause}, \textit{False Causality}, \textit{Causal Oversimplification} $\rightarrow$ \textit{Causal Oversimplification}). We also rephrase some fallacy types by removing words such as ``Appeal to'' (e.g,. \textit{Appeal to Emotion} $\rightarrow$ \textit{Emotional Language}) that tend to throw off generative models causing over prediction of these types as observed in our initial experiments. Some fallacies have partial or full overlap with others across the four schemes. Therefore, we merge these types and use the label of the most frequent or the most representative label of the fallacy type (e.g., \textit{Fallacy of Relevance} $\rightarrow$ \textit{Red Herring}). We also unify the definitions of fallacy types in prompts across datasets. We end up with 28 unique fallacy types across five datasets  
\textsc{Argotario}: 5, \textsc{Propaganda}: 15, \textsc{Logic}: 13, \textsc{Covid-19} and \textsc{Climate}: 9. Complete list of fallacy labels and definitions for all types is shown in Appendix \ref{app:fal}.


\section{Multitask Instruction-based Prompting}
\label{sec:method}
Recently, \citet{wei2021finetuned,sanh2022multitask} leverage the intuition that NLP tasks can be described via natural language instructions, such as ``\textit{Is the sentiment of this movie review positive or negative?}'' or ``\textit{Translate `how are you' into Chinese.}''. They then take a pre-trained language model and perform \textit{instruction tuning} — fine-tuning the model on several NLP datasets expressed via natural language instructions. Such an approach has several benefits, with the most important one being able to have a unified model 
for several tasks. Finally, training tasks spanning 
diverse datasets in a massively multitask fashion improves inference time performance especially for smaller datasets.

Following the success of multitask instruction-based prompting 
we approach different formulations of fallacies across datasets as different tasks with a generic prompting framework in a single model. We use T5 \cite{raffel2020exploring} as the backbone model for training on all five fallacy datasets that have different number and types of fallacies. We hypothesize that when a model is able to learn to recognize fallacy types from multiple datasets, it is more likely able to learn generic traits of fallacy types rather than learning characteristics specific to a single dataset.

A sample list of instructions for each dataset is shown in Figure \ref{fig:model}  (Full list in Appendix \ref{app:prompts}). All instructions start with an n-gram (e.g. \textit{`Given a text segment'}) followed by a list of fallacy types with or without their definitions. The complete set of fallacies and definitions are shown in Appendix \ref{app:fal}. The final component of the instruction is specific to each dataset (e.g., question-answer pair for \textsc{Argotario}, sentence-fragment or sentence only for \textsc{Propaganda}). The generation target during training and test is one of the fallacies types that are permissible for each dataset. In addition, we ask the model to generate the 
fragment that contains the fallacy
(\textsc{Propaganda} dataset only) during training to increase the diversity of prompts and instructions during training. Since the overall objective of this work is to have a generic classifier for fallacy and to compare with other classification methods, evaluating the model's ability to correctly generate the fallacious fragment is beyond the scope of this paper. During inference time, we use greedy decoding and select the generated target as the prediction of fallacy type. The evaluation is done using strict string match with the gold fallacy. Model hyperparameters are shown in Appendix \ref{app:hyp}.



\section{Evaluation Setup and Results}
Given the high imbalance nature of all fallacy datasets, we report both \textit{accuracy} (equivalent to micro F1 as we do not include multi-label instances) and \textit{Macro F1}.

\paragraph{Baselines.}
We consider the following three models as our baselines: i) zero-shot classification using UnifiedQA \cite{khashabi-etal-2020-unifiedqa}; ii) few-shot instruction-tuning of GPT-3 \cite{brown2020language}; and iii) full-shot fine-tuning of BERT \cite{devlin-etal-2019-bert}. UnifiedQA is a question-answering model that is trained on 20 question-answering datasets in different formats and showed generalization capability to unseen data. We use its recent version UnifiedQA-v2 (3B size) \cite{khashabi2022unifiedqa} to test the ability of such model to recognize fallacies in zero-shot settings. 
We also do few-shot instruction-tuning of GPT-3 as many fallacy datasets are of small size, which poses the need for models that can perform well using few-shot training. We setup the instructions in a similar fashion to the ones used for T5 (i.e. \textit{List} prompt in Figure \ref{fig:model}). Additionally, we setup instructions with explanations where each few-shot example has a text segment, a fallacy label, and a sentence explaining why the fallacy label is suitable for the text, which is shown to improve the results of few-shot learning \cite{lampinen2022can}.\footnote{Example instructions shown in Appendix \ref{app:prompts}} Constrained by the length allowed in the prompt, we use 2-shots per the five fallacy types for the \textsc{Argotario} dataset, and 1-shot per the nine-to-fifteen fallacy types for the other datasets. Given the high number of fallacy types, it is not feasible to instruction-tune GPT-3 on the 28 unique fallacy types that exist in all five datasets combined. Finally, we fine-tune BERT for 3 epochs on each dataset separately to test its ability to do fallacy recognition. All model hyperparameter details are shown in Appendix \ref{app:hyp}. 

We also use a T5-large model trained on each dataset separately using the instructions shown in Figure \ref{fig:model} as a baseline in order to compare the results of single-dataset with multi-dataset training.

\begin{table*}[]
        \centering
        \scalebox{0.97}{
        \begin{tabular}{|cll |cc| cc| cc| cc| cc|}
        \hline
        Training &Shot &Model &\multicolumn{2}{c|}{Argotario} &\multicolumn{2}{c|}{Propaganda} &\multicolumn{2}{c|}{Logic} &\multicolumn{2}{c|}{Covid-19} &\multicolumn{2}{c|}{Climate} \\
        Data& & &Acc. &F1 &Acc. &F1 &Acc. &F1 &Acc. &F1 &Acc. &F1\\
        \hline
        -- &Zero &UnifiedQA &23 &14  &04 &01  &21 &08  &14 &07  &08 &02\\
        \hline
        Single &Few &GPT-3  &45 &39 &19 &13 &20 &22 &14 &09 &11 &04 \\
        &Few+Exp &GPT-3 &47 &39 &13 &10 &19 &22 &10 &08 &10 &03 \\
        &Full &BERT &44 &38 &50 &25 &35 &31 &25 &08 &23 &04\\
        &Full &T5-Large &25 &14 &66 &30 &56 &45 &26 &09 &23 &04 \\
        \hline
        Multi &Full &T5-Large &\underline{59} &\underline{59} &\underline{70} &\underline{41} &\underline{68} &\underline{62} &\textbf{31} &\underline{26} &\textbf{27} &\underline{17} \\
        &Full &T5-3B &\textbf{64} &\textbf{64} &\textbf{73} &\textbf{56} &\textbf{70} &\textbf{66} &\underline{29} &\textbf{28} &\underline{25} &\textbf{20} \\
        \hline 
        \end{tabular}
        }
        \caption{Accuracy and Macro F1 scores on all datasets. \textbf{Exp}: explanations added to the few shot examples. Numbers in \textbf{Bold} represents the best score for each dataset, and \underline{underlined} numbers are the second best.}
        \label{tab:res}
\end{table*}
\begin{table*}[]
    \centering
    \scalebox{0.92}{
        \begin{tabular}{|l|c|c c c|}
        \hline
        Model &T5-L &\multicolumn{3}{c|}{T5-3B}\\  \hline
        Prompt &All &All &Def &List\\ \hline
        \small Black-and-White Fallacy &35 &32 &21 &28 \\
        \small Causal Oversimplification &24 &48 &24 &24 \\
        \small Doubt &66 &69 &61 &60 \\
        \small Exaggeration or Minimization &51 &61 &42 &37 \\
        \small Fear or Prejudice &44 &56 &45 &44 \\
        \small Flag-Waving &58 &71 &73 &66 \\
        \small Irrelevant Authority &52 &49 &26 &36 \\
        \small Loaded Language &82 &82 &80 &79 \\
        \small Name Calling or Labeling &83 &83 &82 &82 \\
        \small Red Herring &0 &50 &18 &0 \\
        \small Reductio Ad Hitlerum &0 &37 &0 &0 \\
        \small Slogans &50 &51 &42 &48 \\
        \small Strawman &0 &11 &0 &0 \\
        \small Thought-Terminating Cliches &29 &44 &38 &24 \\
        \small Whataboutism &0 &44 &43 &17 \\
        \hline
        Accuracy &70 &73 &69 &67 \\
        Macro F1 &41 &56 &43 &39 \\
        \hline 
        \multicolumn{5}{c}{(a) Propaganda} \vspace{.45cm}\\ 
        \hline
        Model &T5-L &\multicolumn{3}{c|}{T5-3B}\\  \hline
        Prompt &All &All &Def &List\\ \hline
        \small Ad Hominem &82 &89 &84 &80 \\
        \small Ad Populum &82 &86 &83 &80  \\
        \small Black-and-White Fallacy &88 &84 &87 &89 \\
        \small Causal Oversimplification &70 &81 &65 &79 \\
        \small Circular Reasoning &59 &77 &73 &71 \\
        \small Deductive Fallacy &53 &53 &42 &46 \\
        \small Emotional Language &71 &68 &60 &57 \\
        \small Equivocation &29 &29 &29 &12 \\
        \small Fallacy of Extension &55 &51 &62 &18 \\
        \small Hasty Generalization &74 &70 &69 &68 \\
        \small Intentional Fallacy &26 &33 &24 &12 \\
        \small Irrelevant Authority &60 &70 &66 &58 \\
        \small Red Herring &60 &61 &56 &47 \\
        \hline
        Accuracy &68 &70 &67 &63 \\
        Macro F1 &62 &66 &62 &55 \\
        \hline 
        \multicolumn{5}{c}{(b) Logic}\\
        \end{tabular}}
        \hspace{0.1cm}
        \scalebox{0.917}{
        \begin{tabular}{|l|c| c c c|}
        \hline
        Model &T5-L &\multicolumn{3}{c|}{T5-3B}\\  \hline
        Prompt &All &All &Def &List\\ \hline
        \small Ad Hominem &68 &68 &68 &59 \\
        \small Emotional Language &67 &68 &68 &67 \\
        \small Hasty Generalization &41 &58 &45 &54 \\
        \small Irrelevant Authority &75 &77 &78 &71 \\
        \small Red Herring &44 &52 &41 &48 \\
        \hline
        Accuracy &59 &64 &60 &59 \\
        Macro F1 &59 &64 &60 &59 \\
        \hline 
        \multicolumn{5}{c}{(c) Argotario} \vspace{.2cm}\\
        \hline
        Model &T5-L &\multicolumn{3}{c|}{T5-3B}\\  \hline
        Prompt &All &All &Def &List\\ \hline
        \small Causal Oversimplification &29 &50 &44 &42\\
        \small Cherry Picking &31 &31 &35 &36\\
        \small Evading the Burden of Proof &47 &36 &27 &41\\
        \small False Analogy &40 &33 &50 &33\\
        \small Hasty Generalization &21 &0 &19 &19\\
        \small Irrelevant Authority &57 &24 &0 &15 \\
        \small Red Herring &0 &19 &0 &0 \\
        \small Strawman &0 &24 &0 &0 \\
        \small Vagueness &8 &31 &21 &0 \\
        \hline
        Accuracy &31 &29 &28 &29 \\
        Macro F1 &26 &28 &22 &21 \\
        \hline 
        \multicolumn{5}{c}{(d) Covid-19} \vspace{.2cm}\\
        \hline
        Model &T5-L &\multicolumn{3}{c|}{T5-3B}\\  \hline
        Prompt &All &All &Def &List\\ \hline
        \small Causal Oversimplification &37 &29 &53 &32 \\
        \small Cherry Picking &39 &41 &43 &41 \\
        \small Evading the Burden of Proof &0 &0 &0 &0 \\
        \small False Analogy &25 &0 &0 &25 \\
        \small Hasty Generalization &0 &0 &0 &0 \\
        \small Irrelevant Authority &30 &27 &25 &25 \\
        \small Red Herring &0 &6 &12 &11 \\
        \small Strawman &0 &46 &0 &25 \\
        \small Vagueness &22 &34 &26 &34 \\
        \hline
        Accuracy &27 &25 &29 &28 \\
        Macro F1 &17 &20 &18 &21 \\
        \hline 
        \multicolumn{5}{c}{(e) Climate}\\ 
        \end{tabular}
    }
        \caption{F1 scores for each fallacy type for two T5 model sizes (T5-Large and T5-3Billion), and for three prompt choices (\textbf{Def}: fallacy definitions in prompt; \textbf{List}: fallacy types listed in prompt; \textbf{All}: both Def and List prompts) to study the effect of model size and prompt choice. All models are trained on all five datasets combined.}
        \label{tab:res-prompt-size}
    \end{table*}

\subsection{Multitask Instruction-based Prompting vs. Baselines}

\paragraph{Baseline Results}
\label{sec:res}
Looking at the results shown in Table \ref{tab:res}, UnifiedQA struggles to have any meaningful results and mostly predicts one or two fallacy types for all examples, which shows the infeasibility for models to perform well in zero-shot settings on a complex task such as fallacy recognition. GPT-3 is able to perform well on \textsc{Argotario}, even when trained with 1-shot per class, but struggles to beat any full-shot model on the other datasets, which highlights the difficulty of this task for few-shot training. Adding the explanations does not improve the performance, which could have been outweighed by the low number of shots per class and high number of fallacy classes. We notice that BERT has an acceptable performance on the \textsc{Argotario} dataset (Acc. 44\% and F1 38\%) that has the lowest number of classes (5 fallacy types), which is also the most balanced dataset compared to the other ones. However, when the number of fallacy classes increases to 9 or more, BERT struggles to have a good performance in any of the two evaluation metrics. 

The T5-large models is also trained on each dataset separately using the instructions shown in Figure \ref{fig:model}. It has a surprisingly low performance on the \textsc{Argotario} dataset (Acc. 25\% and F1 14\%) that is significantly lower than BERT and GPT-3. However, it is able to learn better for datasets with high number of classes (13-15 class) and large training data (e.g. \textsc{Propaganda} and \textsc{Logic}).

\paragraph{Multitask Instruction-based Prompting Results} 
We train two sizes of the T5 models (large and 3B) on all datasets combined using the instructions mentioned in Figure \ref{fig:model}. This increases the performance significantly on all datasets of the T5-large model compared to its performance when trained on one dataset at a time as shown in Table \ref{tab:res}. The numbers further improve when we increase the size of the model from T5-large to T5-3B. This shows the benefit of our unified  model based on multitask instruction-based prompting (multi-dataset) for fallacy recognition where we have limited resources and some very small datasets, and also shows the ability of larger models to generalize to the five test sets. The two multi-dataset models always have the best or second best results on all datasets. Also, the T5-3B model is better than T5-large in all accuracy and F1 scores for all datasets excepts accuracy scores for the \textsc{Covid-19} and \textsc{Climate} where the T5-large is better which could be due to having more correct predictions in the majority classes as the T5-3B is still better in macro F1 scores. To further understand the effect of the model size and prompt choice, we discuss in the next section the per-class performance of four different T5 models.

\subsection{Performance of our Unified Model on Fallacy Types}
\label{ssec:fal_type}
We show the per-class (fallacy type) results of our unified model (multitask instruction-based prompting) using two model sizes (T5-large and T5-3B) and three prompts choices (Def, List, and All) in Tables \ref{tab:res-prompt-size}-a to \ref{tab:res-prompt-size}-e.

\paragraph{Model Size}
In general, increasing the model size (from T5-large to T5-3B both trained on all prompts) improves the overall results (especially macro F1) on all datasets. We notice the importance of model size in most datasets for fallacies types that have diversion moves (e.g. \textit{Red Herring} in all datasets, \textit{Strawman} in \textsc{Covid-19} and \textsc{Climate}, \textit{Whataboutism} in \textsc{Propaganda}) where additional context is usually needed to make accurate predictions. A model with more parameters is in principle better in capturing more information during pretraining, which could be more useful for such fallacies that require more information beyond the provided segment. However, this is not always true for other fallacies of diversion. The results are the same or marginally different for the two model sizes when the fallacy is among the majority training classes (e.g. \textit{Cherry Picking} in \textsc{Covid-19} and \textsc{Climate}), or inconsistent due to different conceptualizations of a single fallacy across datasets (e.g. \textit{Irrelevant Authority}, more discussion at the end of this Section). Interestingly, the smaller size model (T5-large) has similar performance to the larger model (T5-3B) on some fallacy types with strong lexical cues contained in the text segment (e.g. \textit{Loaded Language}, \textit{Name Calling} and \textit{Slogans} in \textsc{Propaganda}; \textit{Ad Hominem} and \textit{Emotional Language} in \textsc{Logic} and \textsc{Argotario}).

\paragraph{Prompt Choice}
We also fix the model size (T5-3B) but change the prompts used for training to see which prompt is more useful for this task. We mainly experiment with two prompts that include either the definitions of all fallacies or only listing the names of all fallacies. In both cases, the prompt starts with an instruction followed by either definitions or fallacy names then ending with the segment that has the fallacious text. Including both prompts for each training instance yields the best results in most cases as we would expect. However, it seems that some fallacies benefit more from including the definitions in the prompt than others. In general, including the definitions (T5-3B-Def) rather than just fallacy names (T5-3B-List) has higher accuracy and macro F1 scores in 4 out of 5 datasets as shown in Table \ref{tab:res-prompt-size} (exceptions are accuracy in \textsc{Covid-19} and F1 in \textsc{Climate}). In particular, it seems that definitions are more useful for fallacies that are closely related to other fallacies in one scheme where the definition helps in further clarifying the difference between the two. For example, in \textsc{Propaganda} (Table \ref{tab:res-prompt-size}-a) \textit{Thought-Terminating Cliches} are defined as ``\textit{words or phrases that offer short, simple and generic solutions to problems}'' which is mostly confused with \textit{Loaded Language} by most models, especially ones not trained with definitions. Also in \textsc{Propaganda}, T5-3B-Def has a much higher score than T5-3B-List on \textit{Whataboutism}, which is ``\textit{a discrediting technique that accuse others of hypocrisy}'' which includes introducing questions about other irrelevant matters. This could have caused models to confuse it with the \textit{Doubt} fallacy.

\paragraph{Fallacy Types Across Datasets}
There are two fallacies that exist in all five datasets (i.e. \textit{Irrelevant Authority} and \textit{Red Herring}) and two other fallacies that exist in four datasets (i.e. \textit{Causal Oversimplification} and \textit{Hasty Generalization}). We closely look at these fallacies to understand the challenges posed by changes in domain, genre, and annotation guidelines.

Consider the results shown in Tables \ref{tab:res-prompt-size} (a-e) for \textit{Irrelevant Authority}, we can notice three observations: i) T5-large is the best in \textsc{Propaganda}, \textsc{Covid-19}, and \textsc{Climate}; ii) T5-3B-All is the best in \textsc{Logic} and marginally second best (to T5-3B-Def) in \textsc{Argotario}; iii) similar to model size, including the definition in the prompt has inconclusive benefit across datasets. This can be mainly attributed to inconsistency in how this fallacy is defined in different schemes as for example it strictly refers to ``\textit{mention of false authority on a given matter}'' in \textsc{Covid-19}, while it additionally includes ``\textit{referral to a valid authority but without supporting evidence}'' in \textsc{Propaganda} (all definitions provided in Appendix \ref{app:fal}). 

Similarly, no single model is consistently better in detecting \textit{Red Herring} across all datasets as shown in Tables \ref{tab:res-prompt-size} (a-e). This, however, is more likely caused by the different format this particular fallacy has in different domains and genres as it consists of shorter phrases in \textsc{Propaganda}, asking irrelevant or misleading questions in \textsc{Climate}, and mentions of irrelevant entities in \textsc{Logic}.

\textit{Causal Oversimplification} has more consistent results as shown in Tables \ref{tab:res-prompt-size} (a,b,d,e) where the T5-3B-All model has the best results in three out of four datasets. This illustrates that while the notion of this fallacy might differ across datasets, it still strongly shares common generic features (e.g. the existence of a causal relation) that make it distinguishable by a single model in different settings. 

Finally, the results for \textit{Hasty Generalization} shown in Tables \ref{tab:res-prompt-size} (b-e) indicate that detecting this fallacy becomes more challenging when other similar fallacies exist in a fallacy scheme (e.g. \textit{Cherry Picking} in \textsc{Covid-19} and \textsc{Climate}), and less challenging when other fallacies in the scheme are further away (e.g. \textsc{Logic} and \textsc{Argotario}).

Nevertheless, this multitask setup provides the model with the opportunity to learn to detect specific fallacy types as they are expressed differently, and grouped with different fallacies, which consistently and significantly improves the overall results of fallacy recognition over single-scheme (or single dataset) models.

\begin{table}[]
    \small
    \centering
    \begin{tabular}{|p{7cm}|}
    \hline
        \vspace{.05cm}
        His opinion is: 
        "She may very well believe everything she’s saying, and that is one of the signs of lunacy, believing something that isn’t real.” And her lawyer is \underline{even loonier}\vspace{.05cm}\\
        \hspace{.5cm}{\footnotesize
        \hlc[Goldenrod]{Doubt}
        \hlc[LimeGreen]{Name Calling or Labeling}
        \hlc[CarnationPink]{Doubt}
        }
        \vspace{.05cm}\\ 
    \hline
        \vspace{.05cm}
        "\underline{Christianity is Europe's last hope}," Orban told an audience of party faithful at the foot of the Royal Castle in Budapest.
        \vspace{.05cm}\\
        \hspace{1cm}{\footnotesize
        \hlc[Goldenrod]{Slogans}
        \hlc[LimeGreen]{Flag-Waving}
        \hlc[CarnationPink]{Flag-Waving}
        }
        \vspace{.05cm}\\ 
    \hline
        \vspace{.05cm}
        "Orban is \underline{openly Christian} and seems to understand something that many do not and that is you do not allow a wholesale flood of antichrists to pour into your country."
        \vspace{.1cm}\\
        {\footnotesize
        \hlc[Goldenrod]{Flag-Waving}
        \hlc[LimeGreen] {Name Calling or Labeling}
        \hlc[CarnationPink]{Flag-Waving}
        }
        \\ 
    \hline
    \end{tabular}
    \caption{Example sentences from \textsc{Propaganda} with \hlc[Goldenrod]{gold label}, \hlc[LimeGreen]{model prediction} and \hlc[CarnationPink]{expert annotation}. \underline{Underlined} text highlights the propagandistic fragment.}
    \label{tab:errors}
\end{table}

\subsection{Error Analysis}
\label{ssec:error}
In order to better understand model errors and quality of annotations for this complex task for both humans and machines, 
an expert looked at 70 wrongly predicted examples from the \textsc{Propaganda} datasets (5 examples each from 14 propaganda technique, \textit{Strawman} was not included due to low counts). First,  
the expert looked only at the sentence and the fragment identified by the gold annotation as containing a fallacy and she independently annotated the propaganda technique at stake.  
Comparing this annotation with gold labels and model prediction (T5-3B-All), it turns out that the expert annotator agreed with the gold label in 75\% of the cases, and with the model prediction in 15\%, while she chose a different label in 10\% of the cases. Table \ref{tab:errors} shows three examples along with 
gold labels, model predictions, and expert annotations.

Consider the first example in Table \ref{tab:errors} that has \textit{Doubt} as the gold label. The expert agrees that the propaganda technique used rests on questioning the credibility of the lawyer (\textit{Doubt}), even though the adjective ``lunatic" is a literal instance of \textit{Name Calling}. Thus, the label predicted by the model is not wrong, but less relevant since it is the lack of trustworthiness the most effective feature in undermining the antagonist's stance, regardless whether it is due to lunacy or lack of integrity. 

In the second example of Table \ref{tab:errors}, the expert agrees with the model prediction of a \textit{Flag-Waving} fallacy in the underlined segment rather than a \textit{Slogan} as the gold label. The term ``\textit{last hope}'' can be considered a slogan, however, when we consider the full propagandistic segment that includes the word ``\textit{Christianity}'', it maps better to \textit{Flag-Waving} as it has been defined in the guidelines (and included in the prompt) as ``\textit{Playing on strong national feeling (or to any group)...}''.

The third example highlights even more the importance of the selected fragment in the prompt: without considering the reference to the ``\textit{antichrist}'' threat, it is not possible to understand that the sentence is playing on a religious-based national feeling. 


Considering the analysis of the 70 examples in the \textsc{Propaganda} dataset, the following general observations are found: i) some fallacious segments can map to more than one fallacy, especially when one of the two is a language fallacy (e.g., \textit{Name Calling, Exaggeration, Loaded Language}). In such cases, the model tends to privilege the language fallacy type, even if usually not the most relevant from an argumentative perspective; ii) for some cases, the expert annotator had to read more context beyond the sentence; 
iii) for some cases, the expert agreed with the gold label but disagreed with the boundaries of the annotated 
fragment by choosing a larger or more informative one. 

In light of this, 
improving automatic fallacy identification may entail i) considering additional context; 
ii) adopting a fallacy scheme with a heuristics that imposes an order into fallacy recognition (structural fallacy followed by diversion and logical fallacies with language fallacies at last when all the others are excluded).

\section{Related Work}
\label{sec:background}
\paragraph{Fallacy} There are various typologies of fallacies that address informal logic traditions or rules of ideal critical discussion  \cite{hansen1996aristotle,van2002argumentation,tindale2007fallacies,walton2008argumentation,damer2012attacking}. This intersects with propaganda techniques that focus on faulty reasoning and emotional appeals to accomplish persuasion \cite{Miller1939propaganda,jowett2012propaganda,torok2015symbiotic,weston2018rulebook}. Computational work on fallacy recognition includes fallacies in dialogue \cite{habernal-etal-2017-argotario,sheng-etal-2021-nice}, argument sufficiency \cite{stab-gurevych-2017-recognizing}, name calling on Reddit \cite{habernal-etal-2018-name}, non-sequitur fallacy in legal text \cite{nakpih2020automated}, logical fallacies \cite{jin2022logical}, fallacies in misinformation \cite{musi2022developing}, as well as propaganda techniques \cite{da-san-martino-etal-2019-fine}. However, all previous work is limited to one fallacy scheme while we develop a model that detects fallacies in five datasets across four fallacy schemes.

\paragraph{Prompting} Using prompts has emerged as a generic framework to train natural language processing models on multiple tasks using prefix text \cite{raffel2020exploring}, and few-shot prompt-tuning of GPT-3 \cite{brown2020language}. This was followed by multiple studies that use prompts on smaller size models using few and full shots on tasks such as natural language inference \cite{schick-schutze-2021-just}, text classification \cite{schick-schutze-2021-exploiting,gao-etal-2021-making},  relation extraction \cite{chen2022knowprompt}, and using instruction prompts for multiple tasks \cite{mishra-etal-2022-cross,sanh2022multitask}. We follow a similar setup by training a T5 model using instruction prompts for different formulations of fallacy recognition approached as multiple tasks.

\section{Conclusion}
\label{sec:conc}
We introduced a unified model using multitask instruction-based prompting for solving the challenges faced by the fallacy recognition task. 
We could unify all the  datasets by converting 28 fallacy types across 5 different datasets into natural language instructions. 
We showed that our unified model 
is better than training on a single dataset. We analyzed the effect of model size and prompt choice 
on the detection of specific fallacy types that could require additional knowledge better captured by bigger models (e.g., diversion fallacies such as \textit{Red Herring}), and the distinction between similar fallacies better detected by more comprehensive prompts that include definitions of fallacy types (e.g., \textit{Doubt} vs. \textit{Whataboutism}). We analyzed the differences of fallacy types that appear in multiple fallacy schemes across the five datasets and showed that one fallacy type could have multiple meanings which further increases the complexity of this task (e.g., \textit{Irrelevant Authority}). We 
conducted a thorough error analysis and 
released a new fallacy dataset for fact-checked content in the climate change domain. 

\section*{Limitations}
\label{sec:limit}

In the current setup, we consider all examples as fallacious or partially fallacious and do not include a "No Fallacy" class, which some of the fallacy datasets have. Based on this assumption, the model's task is to detect the type of fallacy given a fallacious example. Including "No Fallacy" makes the datasets severely imbalanced (e.g. 70\% of \textsc{Propaganda} and 50\% of \textsc{Covid-19} are labeled as "No Fallacy"). We elected to remove it for this work since not all datasets have a "No Fallacy" class (e.g. \textsc{Logic}) and since this class is bigger than all 28 fallacy class combined. Even with our initial experiments with downsampling of "No Fallacy" using BERT, the results were not promising. This setup is in line with the propaganda technique classification task \cite{da-san-martino-etal-2020-semeval} and the logical fallacy detection task \cite{jin2022logical} that all do not include "No Fallacy" class. We leave further experimentation of pipeline or joined approaches to separate fallacies from non-fallacies text for future work. Other limitations include the need for external knowledge and the multi-labeling nature of some examples as discussed in Section \ref{ssec:error}, which we leave for future work.

We experiment with the second and third largest sizes of the T5 model, T5-3B (11GB) and T5-large (3GB) and do not run experiments with T5-11B  (40GB) due to lack of resources. The T5-3B is run on 2 Nvidia A-100 GPUs with 40GB memory each with a batch size of 2. These GPU requirements could pose a limitation on using such models in resource-poor settings. They could also have environmental impacts if trained (and re-trained) for longer periods of time. The training time of the T5-3B on 2 GPUs for 5 epochs is on average 2-3 hours depending on the size of the dataset.

\section*{Ethics Statement}
\label{sec:ethics}
The intended use of our fallacy recognition model is as a human-in-the-loop assistant to signal misinformation or  
for teaching humans how to identify and avoid using fallacies.  

\section*{Acknowledgments}
\label{sec:acknowledgments}
The first author is grateful for the KACST Graduate Studies Scholarship that supported his PhD research for five years.

\bibliography{anthology,custom}
\bibliographystyle{acl_natbib}

\newpage
\appendix
\section{Model Hyperparameters}
\label{app:hyp}
We use huggingface's implementation \cite{wolf-etal-2020-transformers} of the T5 model (large and 3B) where we train all models for 5 epochs choosing the epoch with lowest evaluation loss as the final model. The models are run with 1e-4 learning rate, Adam optimizer, batch size 2, gradient accumulation steps 512, maximum source length 1024, maximum target length 64. At inference time, the target is generated using greedy decoding (beam search of size 1) with no sampling and default settings for T5. The generated target is then compared with the fallacies in the given scheme and the prediction is counted as correct if they are the same using strict string match.

We also use huggingface's implementation of BERT (base) and fine-tune the model for 3 epochs with 1e-5 learning rate, batch size 16, maximum sequence length 256. 

For GPT-3, we use the completion API of OpenAI \cite{brown2020language} using their large engine that is trained with instructions (\textit{text-davinci-002}) with temperature 0, max generated tokens 150 and other parameters kept at default value (e.g. top\_p=1). The generated target is considered correct if it has the gold fallacy (even with additional text). Since GPT-3 is trained with few-shots only, it sometimes generates some generic prefix, repeats the text segment, or generates more than one fallacy.

\begin{table*}[]
    \small
    \centering
    \begin{tabular}{|l |p{3.25cm} | p{10.75cm}|}
    \hline
    &Fallacy Type &Definition\\
    \hline \hline
        \multirow{10}{*}{\STAB{\rotatebox[origin=c]{90}{\cite{habernal-etal-2017-argotario}}}}
        &Ad Hominem &\textbf{The opponent attacks a person instead of arguing against the claims that the person has put forward.} \\ \cline{2-3}
        &Appeal to Emotion &\textbf{This fallacy tries to arouse non-rational sentiments within the intended audience}  \\
        &(Emotional Language) &\textbf{in order to persuade.} \\ \cline{2-3}
        &Hasty Generalization &The argument uses a sample which is too small, or follows falsely from a sub-part to a composite or the other way round. \\ \cline{2-3}
        &Irrelevant Authority &While the use of authorities in argumentative discourse is not fallacious inherently, appealing to authority can be fallacious if the authority is irrelevant to the discussed subject. \\ \cline{2-3}
        &Red Herring &This argument distracts attention to irrelevant issues away from the thesis which is supposed to be discussed. \\ \hline
    \hline
        \multirow{35}{*}{\STAB{\rotatebox[origin=c]{90}{\cite{da-san-martino-etal-2019-fine}}}} &Black and White Fallacy &\textbf{Presenting two alternative options as the only possibilities, when in fact more possibilities exist. As an the extreme case, tell the audience exactly what actions to take, eliminating any other possible choices (Dictatorship).} \\ \cline{2-3}
        &Causal Oversimplification &\textbf{Assuming a single cause or reason when there are actually multiple causes for an issue.} \\ \cline{2-3}
        &Doubt &\textbf{Questioning the credibility of someone or something.}   \\ \cline{2-3}
        &Exaggeration &\textbf{Either representing something in an excessive manner: making things larger,} \\ 
        &or Minimization &\textbf{better, worse or making something seem less important than it really is} \\ \cline{2-3}
        &Appeal to fear/prejudice  &\textbf{Seeking to build support for an idea by instilling anxiety and/or panic in the} \\
        &(Fear or Prejudice) &\textbf{population towards an alternative. In some cases the support is based on preconceived judgements.} \\ \cline{2-3}
        &Flag-Waving &\textbf{Playing on strong national feeling (or to any group) to justify/promote an action/idea.} \\ \cline{2-3}
        &Appeal to Authority &Stating that a claim is true simply because a valid authority or expert on the issue said\\
        &(Irrelevant Authority) &it was true, without any other supporting evidence offered. We consider the special case in which the reference is not an authority or an expert in this technique, although it is referred to as Testimonial in literature.  \\ \cline{2-3}
        &Loaded Language &\textbf{Using specific words and phrases with strong emotional implications (either positive or negative) to influence an audience.} \\ \cline{2-3}
        &Name Calling or Labeling &\textbf{Labeling the object of the propaganda campaign as either something the target audience fears, hates, finds undesirable or loves, praises.} \\ \cline{2-3}
        &Red Herring &Introducing irrelevant material to the issue being discussed, so that everyone's attention is diverted away from the points made. \\ \cline{2-3}
        &Reductio Ad Hitlerum &\textbf{Persuading an audience to disapprove an action or idea by suggesting that the idea is popular with groups hated in contempt by the target audience. It can refer to any person or concept with a negative connotation.} \\ \cline{2-3}
        &Slogans &\textbf{A brief and striking phrase that may include labeling and stereotyping. Slogans tend to act as emotional appeals.}  \\ \cline{2-3}
        &Strawman &\textbf{When an opponent's proposition is substituted with a similar one which is then refuted in place of the original proposition.} \\ \cline{2-3}
        &Thought-Terminating &\textbf{Words or phrases that discourage critical thought and meaningful discussion}\\
        &Cliches &\textbf{about a given topic. They are typically short, generic sentences that offer seemingly simple answers to complex questions or distract attention away from other lines of thought.} \\ \cline{2-3}
        &Whataboutism &\textbf{A technique that attempts to discredit an opponent's position by charging them with hypocrisy without directly disproving their argument.} \\ \hline
    \hline 
        \multirow{25}{*}{\STAB{\rotatebox[origin=c]{90}{\cite{jin2022logical}}}} 
        &Ad Hominem &An irrelevant attack towards the person or some aspect of the person who is making the argument, instead of addressing the argument or position directly. \\ \cline{2-3}
        &Ad Populum &\textbf{A fallacious argument which is based on affirming that something is real or better because the majority thinks so.} \\ \cline{2-3}
        &False Dilemma &A claim presenting only two options or sides when there are many options or sides. \\
        &(Black and White Fallacy) & \\ \cline{2-3}
        &False Causality  (Causal &A statement that jumps to a conclusion implying a causal relationship without \\
        &Oversimplification) &supporting evidence \\ \cline{2-3}
        &Circular Reasoning &\textbf{A fallacy where the end of an argument comes back to the beginning without having proven itself.} \\ \cline{2-3}
        &Deductive Fallacy & \textbf{An error in the logical structure of an argument.} \\ \cline{2-3}
        &Appeal to Emotion &Manipulation of the recipient’s emotions in order to win an argument. \\
        &(Emotional Language) & \\ \cline{2-3}
        &Equivocation &\textbf{An argument which uses a phrase in an ambiguous way, with one meaning in one portion of the argument and then another meaning in another portion.} \\ \cline{2-3}
        &Fallacy of Extension &\textbf{An argument that attacks an exaggerated/caricatured version of an opponent’s.} \\ \cline{2-3}
    \end{tabular}
    \caption{Fallacy Names and Definitions (\textbf{Bold}: definition of this fallacy used in all prompts for across datasets)}
    \label{tab:fal_def_label}
\end{table*}

\begin{table*}[]
    \small
    \centering
    \begin{tabular}{|l |p{3.25cm} | p{10.75cm}|}
    \cline{2-3}
        &Faulty Generalization &An informal fallacy wherein a conclusion is drawn about all or many instances of a \\
        &(Hasty Generalization) &phenomenon on the basis of one or a few instances of that phenomenon is an example of jumping to conclusions. \\ \cline{2-3}
        &Intentional Fallacy &\textbf{Some intentional/subconscious action/choice to incorrectly support an argument.} \\ \cline{2-3}
        &Fallacy of Credibility &An appeal is made to some form of ethics, authority, or credibility. \\
        &(Irrelevant Authority) & \\ \cline{2-3}
        &Fallacy of Relevance &Also known as red herring, this fallacy occurs when the speaker attempts to divert \\
        &(Red Herring) &attention from the primary argument by offering a point that does not suffice as counterpoint/supporting evidence (even if it is true). \\ \hline
        \hline
        \multirow{13}{*}{\STAB{\rotatebox[origin=c]{90}{\cite{musi2022developing}}}} 
        &Evading Burden of Proof &\textbf{A position is advanced without any support as if it was self-evident.} \\ \cline{2-3}
        &Cherry Picking &\textbf{The act of choosing among competing evidence that which supports a given position, ignoring or dismissing findings which do not support it.} \\ \cline{2-3}
        &Red Herring &\textbf{The argument supporting the claim diverges the attention to issues which are irrelevant for the claim at hand.} \\ \cline{2-3}
        &Strawman &The arguer misinterprets an opponent's argument for the purpose of more easily attacking it, demolishes the misinterpreted argument, and then proceeds to conclude that the opponent's real argument has been demolished. \\ \cline{2-3}
        &False Authority &\textbf{An appeal to authority is made where the it lacks credibility or knowledge in the}\\
        &(Irrelevant Authority) &\textbf{discussed matter or the authority is attributed a tweaked statement.} \\ \cline{2-3}
        &Hasty Generalization &\textbf{A generalization is drawn from a sample which is too small, not representative or not applicable to the situation if all the variables are taken into account.} \\ \cline{2-3}
        &False Cause (Causal&X is identified as the cause of Y when another factor Z causes both X and Y \\ 
        & Oversimplification) &OR X is considered the cause of Y when actually it is the opposite   \\ \cline{2-3}
        &Post Hoc (Causal&It is assumed that because B happens after A, it happens because of A. In other words \\
        & Oversimplification) &a causal relation is attributed where, instead, a simple correlation is at stake \\ \cline{2-3}
        &False Analogy &\textbf{because two things [or situations] are alike in one or more respects, they are necessarily alike in some other respect.} \\ \cline{2-3}
        &Vagueness &\textbf{A word/a concept or a sentence structure which are ambiguous are shifted in meaning in the process of arguing or are left vague being potentially subject to skewed interpretations.} \\
    \hline
    \end{tabular}
    \caption{ ({\it continue}) Fallacy Names and Definitions (\textbf{Bold}: definition of fallacy used in all prompts across datasets)}
    \label{tab:fal_def_label_2}
\end{table*}

\begin{table*}[]
    \centering
    \scalebox{.74}{
    \begin{tabular}{| ll| | c|c|c|c|c ||c|c|}
    \hline
    &\multirow{2}{*}{Fallacy} &\textsc{Argotario} &\textsc{Propaganda} &\textsc{Logic} &\textsc{Covid-19} &\textsc{Climate} &\underline{Total} &\underline{Total}\\
    & &train/dev/test &train/dev/test &train/dev/test &train/dev/test &train/dev/test &train/dev/test &All\\
    \hline \hline
        1 &Ad Hominem  &102 /26/ 31 &--/--/-- &406 /64/ 81 &--/--/-- &--/--/-- &508 /90/ 112 &710\\
        2 &Ad Populum &--/--/-- &--/--/-- &296 /81/ 62 &--/--/-- &--/--/-- &296 /81/ 62 &439\\
        3 &B\&W Fallacy &--/--/-- &60 /16/ 19 &192 /40/ 25 &--/--/-- &--/--/-- &252 /56/ 44 &352\\
        4 &Causal Ov.simp. &--/--/-- &111 /28/ 34 &303 /49/ 36 &36 /10/ 10 &39 /10/ 11 &489 /97/ 91 &677\\
    \hline
        5 &Cherry Picking &--/--/-- &--/--/-- &--/--/-- &76 /20/ 23 &67 /17/ 21 &143 /37/ 44 &224\\
        6 &Circular Reason. &--/--/-- &--/--/-- &238 /40/ 35 &--/--/-- &--/--/-- &238 /40/ 35  &313\\
        7 &Deductive &--/--/-- &--/--/-- &205 /28/ 31 &--/--/-- &--/--/-- &205 /28/ 31  &264\\
        8 &Doubt &--/--/-- &263 /66/ 82 &--/--/-- &--/--/-- &--/--/-- &263 /66/ 82  &411\\
    \hline
        9 &Emotional Lang. &150 /38/ 47 &--/--/-- &230 /38/ 41 &--/--/-- &--/--/-- &380 /76/ 88  &544\\
        10 &Equivocation &--/--/-- &--/--/-- &62 /13/ 11 &--/--/-- &--/--/-- &62 /13/ 11  &86\\
        11 &Evad Burd Prf &--/--/-- &--/--/-- &--/--/-- &76 /20/ 23 &31 /8/ 9 &107 /28/ 32  &167\\
        12 &Exag/Mini &--/--/-- &304 /76/ 94 &--/--/-- &--/--/-- &--/--/-- &304 /76/ 94  &474\\
    \hline
        13 &Extension &--/--/-- &--/--/-- &187 /31/ 46 &--/--/-- &--/--/-- &187 /31/ 46  &264\\
        14 &False Analogy &--/--/-- &--/--/-- &--/--/-- &13 /5/ 3  &17 /5/ 5 &30 /10/ 8  &48\\
        15 &Fear/Prejudice &--/--/-- &131 /33/ 41 &--/--/-- &--/--/-- &--/--/-- &131 /33/ 41 &205\\
        16 &Flag-Waving &--/--/-- &145 /37/ 45 &--/--/-- &--/--/-- &--/--/-- &145 /37/ 45  &227\\
    \hline
        17 &Hasty General. &104 /26/ 32 &--/--/-- &561 /128/ 123 &54 /15/ 16 &4 /2/ 2 &723 /171/ 173  &1,067\\
        18 &Intentional Fal. &--/--/-- &--/--/-- &215 /34/ 26 &--/--/-- &--/--/-- &215 /34/ 26 &275\\
        19 &Irrelevant Auth. &92 /24/ 29 &57 /15/ 17 &196 /18/ 33 &26 /8/ 8 &32 /8/ 10 &403 /73/ 97  &573\\
        20 &Loaded Lang. &--/--/-- &1,331/333/416 &--/--/-- &--/--/-- &--/--/-- &1,331/333/416 &2,080  \\
    \hline
        21 &Name Calling &--/--/-- &685 /172/ 214 &--/--/-- &--/--/-- &--/--/-- &685 /172/ 214 &1,071\\
        22 &Red Herring &115 /29/ 35 &16 /4/ 10 &214 /43/ 46 &28 /8/ 8 &44 /12/ 13 &417 /96/ 112  &625\\
        23 &Reductio AH. &--/--/-- &33 /9/ 10 &--/--/-- &--/--/-- &--/--/-- &33 /9/ 10 &52\\
        24 &Slogans &--/--/-- &84 /22/ 26 &--/--/-- &--/--/-- &--/--/-- &84 /22/ 26 &132 \\
    \hline
        25 &Strawman &--/--/-- &4/1/6 &--/--/-- &28 /8/ 8 &23 /6/ 7 &55 /15/ 21 &91\\
        26 &Thought-Term. &--/--/-- &48 /12/ 14 &--/--/-- &--/--/-- &--/--/-- &48 /12/ 14 &74 \\
        27 &Vagueness &--/--/-- &--/--/-- &--/--/-- &53 /15/ 23 &48 /12/ 14 &101 /27/ 37  &165\\
        28 &Whataboutism &--/--/-- &33 /9/ 10 &--/--/-- &--/--/-- &--/--/-- &33 /9/ 10 &52\\
    \hline \hline
        &Total (tr/de/te)
        &563 /143/ 174 &3,305 /833/ 1,038 &3,305 /607/ 596 &390 /109/ 122 &305 /80/ 92 &\multicolumn{2}{c|}{7,868 /1,772/ 2,022}\\ \cline{8-9}
        &Total (All)
        &880 &5,176 &4,508 &621 &477 &\multicolumn{2}{c|}{11,662}\\
    \hline 
    \end{tabular}}
    \caption{Counts of fallacy types in each split across all datasets.}
    \label{tab:counts}
\end{table*}



\section{Fallacy Datasets}
\label{app:fal}
We list in Tables \ref{tab:fal_def_label} and \ref{tab:fal_def_label_2} all the definitions and fallacy labels used in our prompts. As mentioned in Section \ref{sec:data}, we unify the definitions and labels for fallacies that fully or partially overlap. Additionally, in the same tables we show the original labels and definitions for all four fallacy schemes as they are released by \cite{habernal-etal-2017-argotario} for \textsc{Argotario}, \cite{da-san-martino-etal-2019-fine} for \textsc{Propaganda}, \cite{jin2022logical} for \textsc{Logic}, and \cite{musi2022developing} for \textsc{Misinformation} that is used by the \textsc{Covid-19} and \textsc{Climate} datasets. We also show counts of fallacy types in training/dev/test splits for all datasets in Table \ref{tab:counts}. Below is a detailed description of the four existing fallacy datasets.

\paragraph{\textsc{Argotario}} Introduced by \citet{habernal-etal-2017-argotario}, the Argotario dataset consists of five fallacies in dialogue between players in game settings. The five fallacy types are: \textit{Ad Hominem, Appeal to Emotion, Red Herring, Hasty Generalization, irrelevant authority}, in addition to the \textit{No Fallacy} type. These types are selected because they are: common in argumentative discourse, distinguishable from each other, and have different difficulty levels. Players in the game are presented with a topic (question), which they answer using one of the fallacy types. Other players then try to predict the fallacy type written by author of the answer. The final label is determined when at least four players agree with the author of the answer on the type of fallacy. Each instance consist of a question-answer pair and one out of five fallacy labels.

\paragraph{\textsc{Propaganda}} \citet{da-san-martino-etal-2019-fine} identified 18 propaganda techniques that appear in news articles. We focus on the following 15 of them that have a fallacy and frequent enough in the data: \textit{Loaded Language, Name Calling or Labeling, Exaggeration or Minimization, Doubt, Appeal to Fear/Prejudice, Flag-Waving, Causal Oversimplification, Slogans, Appeal to Authority, Black-and-White Fallacy, Thought-Terminating Cliche, Whataboutism, Reductio ad Hitlerum, Red Herring,} and \textit{Strawman}. We ignore propaganda techniques that do not have an argumentative fallacy (e.g. \textit{Repetition}) or not frequent enough in the data (e.g. \textit{Bandwagon, OIVC}). The authors annotate the text spans and propaganda technique (fallacy type) in 451 articles from 48 news outlets allowing multiple labels and partial overlap of text spans. We frame this at the sentence level where the fallacy type becomes the label of the sentence if the fragment is included within the sentence. For sentences with multiple fragments, we consider the label of the longer fragment. We ignore propaganda fragments that span across multiple sentences. This is the biggest dataset in our experiments but it is also the most imbalance one where 6 out the 18 propaganda techniques represent more than 80\% of all propagandistic segments. Each training instance consists of a sentence, a fragment, and one out of fifteen fallacy labels.

\paragraph{\textsc{Logic}} \citet{jin2022logical} collected examples of logical fallacies from educational websites on fallacies such as Quizziz, study.com and ProProfs. They identified 13 types of fallacies in the dataset using Wikipedia\footnote{\href{https://en.wikipedia.org/wiki/List\_of\_fallacies}{https://en.wikipedia.org/wiki/List\_of\_fallacies}} as a reference. The fallacy types are: \textit{Faulty Generalization, False Causality, Circular Claim, Ad Populum, Ad Hominem, Deductive Fallacy, Appeal to Emotion, False Dilemma, Equivocation, Fallacy of Extension, Fallacy of Relevance, Fallacy of Credibility} and \textit{Intentional Fallacy}. Each training instance consists of a text segment (e.g. dialogue, sentence) and one of thirteen fallacy labels. The authors also introduce another challenge dataset: \textsc{ClimateLogic} that follows the same fallacy scheme. However, it contains text segments that are too long (e.g. multiple paragraphs) with no annotations of smaller fallacious fragments like the Propaganda dataset. Therefore, \textsc{ClimateLogic} is beyond the scope of this study.

\paragraph{Misinformation} \citet{musi2022developing} identified 10 fallacies though analysis of fact-checked news (article and social media posts) about \textsc{Covid-19}. They consider fallacies as indicators of misinformation, which they define as misleading news that is not necessarily false communicated with the intention to deceive, thus making it harder to detect and fact-check. The fallacies are: \textit{Structural (Evading the Burden of Proof), Diversion (Cherry Picking, Strawman, Red Herring, False Authority), Logical (Hasty Generalization, Post Hoc, False Cause, False Analogy), and Language (Vagueness)}. They annotate 1,135 covid-19 news and social media posts (621 fallacious) that are fact-checked by five fact-checking organizations.

\begin{table}[t]
    \small
    \centering
    \begin{tabular}{|p{7cm}|}
    \hline
\\
Given the question and answer pairs below, which of the following fallacies occur in the answers: \textit{Emotional Language, Red Herring, Hasty Generalization, Ad Hominem}, or \textit{Irrelevant Authority}?
\\\hspace{1cm}----------------------------------------------------
\\1) {\bf Question:} Is Christianity a peaceful religion?
\\{\bf Answer:} You are the antichrist, you want to destroy our belief in god.
\\{\bf Fallacy:} Ad Hominem
\\{\bf Explanation:} It is an ad hominem becase the speaker is attacked for his bad intentions and not for the point she is making.
\\
\\2) {\bf Question:} Is television an effective tool in building the minds of children?
\\{\bf Answer:} All TV-Shows are bad. Look at "the bachelor". Children cannot learn from it.
\\{\bf Fallacy:} Hasty Generalization
\\{\bf Explanation:} It is a hasty generalization since the evaluation of a whole category is drawn from the evaluation of a single element of the category. 
\\
\\ ...
\\
\\5) {\bf Question:} Should we allow animal testing for medical purposes?
\\{\bf Answer:} No, animals are so cuuuuteeeeeeee!!!
\\{\bf Fallacy:} Emotional Language
\\{\bf Explanation:} It is a fallacy of emotional language since the argument appels to positive emotions associated to animals' appearances.
\\
\\\hlc[YellowGreen]{6) Question: Should gorillas be held in zoos
\\Answer: No, I don't like gorillas.}
\\\hspace{1cm}----------------------------------------------------
\\\hlc[Goldenrod]{Fallacy: Red Herring}
\\ \\
    \hline
    \end{tabular}
    \caption{Example of GPT-3 few-shot instruction with explanations.\hlc[YellowGreen]{Test Example} \hlc[Goldenrod]{Generated Fallacy Type}}
    \label{tab:explanation}
\end{table}
\begin{table*}[t!]
        \scalebox{.88}{
        \centering
        \begin{tabular}{|l|l|l|l|}
        \hline
        Name &Data &Instruction &Target\\
        \hline
        List &\textsc{Argotario} &Given the question and answer below, which of the following &\{fallacy\}\\ 
                        & & fallacies occur in the answer: \{fallacies\}? & \\
                        &&Question: \{question\} Answer: \{answer\} &\\
        \hline
        Def &\textsc{Argotario} &Given the question, answer and definitions, determine which &\{fallacy\}\\ 
                && of the fallacies defined below occur in the answer. &\\
                &&Definitions: \{definitions\} Question: \{question\} Answer: \{answer\} &\\
        \hline \hline
        List &\textsc{Propaganda} &Given the sentence and fragment below, which of the following &\{fallacy\}\\ 
                        && fallacies occur in the fragment: \{fallacies\}?  &\\
                        && Sentence: \{sentence\} Fragment: \{fragment\} &\\
        \hline
        List &\textsc{Propaganda} &Given the sentence below, which of the following fallacies&\{fallacy\}\\ 
                    &(no fragment) & does it have: \{fallacies\}?  &\\
                    && Sentence: \{sentence\} &\\
        \hline
        List &\textsc{Propaganda}   &Given the sentence below, detect the fallacious fragment in it, and the&\{fragment\}\\ 
        +Frag       &&type of fallacy of the fragment from the following: \{fallacies\}.   & \{fallacy\}  \\
                    && Sentence: \{sentence\} &\\
        \hline
        Def &\textsc{Propaganda} &Given the following sentence, fragment and definitions, determine which &\{fallacy\}\\ 
                        && of the fallacies defined below occurs in the fragment: \{fallacies\}.  &\\
                        && Definitions: \{definitions\} &\\
                        &&Sentence: \{sentence\} Fragment: \{fragment\} &\\
        \hline
        Def &\textsc{Propaganda} &Given the sentence and definitions below, determine which &\{fallacy\}\\ 
                    &(no fragment)& of the fallacies defined below occurs in the sentence: \{fallacies\}.  &\\
                    && Definitions: \{definitions\} Sentence: \{sentence\}&\\
        \hline
        Def &\textsc{Propaganda}    &Given the following sentence and definitions, detect the fallacious & \{fragment\}\\ 
        +Frag       &&fragment in the sentence, and the fallacy type of that fragment from &\{fallacy\}  \\
                    && the ones defined below. &\\
                    && Definitions: \{definitions\} Sentence: \{sentence\} &\\
        \hline \hline
        List &\textsc{Logic} &Given the segment below, which of the following &\{fallacy\}\\ 
                        &&fallacies does it have: \{fallacies\}?&\\
                        &&Segment: \{segment\} &\\
        \hline
        Def &\textsc{Logic} &Given the following segment and definitions, determine which &\{fallacy\}\\ 
                        && of the fallacies defined below occurs in the segment. &\\
                        && Definitions: \{definitions\} Segment: \{segment\} &\\
        \hline \hline
        List &\textsc{Covid-19} &Given the segment below, which of the following &\{fallacy\}\\ 
                        &&fallacies does it have: \{fallacies\}?&\\
                        &&Segment: \{segment\} &\\
        \hline
        Def &\textsc{Covid-19} &Given the following segment and definitions, determine which &\{fallacy\}\\ 
                        && of the fallacies defined below occurs in the segment. &\\
                        && Definitions: \{definitions\} Segment: \{segment\} &\\
        \hline \hline
        List &\textsc{Climate} &Given the segment below, which of the following&\{fallacy\}\\ 
                       && fallacies does it have: \{fallacies\}? &\\
                        && Segment: \{segment\} &\\
        \hline
        Def &\textsc{Climate} &Given the segment and definitions below, determine which &\{fallacy\}\\ 
                       && of the fallacies defined below occurs in the segment. &\\
                        &&Definitions: \{definitions\} Segment: \{segment\}&\\
        \hline             
        List &\textsc{Climate} &Given the segment and comment below, which of the &\{fallacy\}\\ 
        +Com               && following fallacies occur in the segment: \{fallacies\}? &\\
                        && Segment: \{segment\} Comment: \{comment\} &\\
        \hline
        Def &\textsc{Climate} &Given the segment, comment and definitions below, determine which &\{fallacy\}\\ 
        +Com                && of the fallacies defined below occurs in the segment. &\\
                        &&Definitions: \{definitions\} Segment: \{segment\} Comment: \{comment\} &\\
        \hline
        \end{tabular}
        }
        \caption{All Instructions. \textbf{Def}: fallacy definition included in the prompt.  \textbf{List}: fallacy types listed in the prompt. \textbf{Frag}: fallacious fragment and fallacy type are the generation targets. \textbf{Com}: fact-checker comments are included, which sometimes explain the fallacy of the segment (\textbf{Frag} and \textbf{Com} are done during training only).}
        \label{tab:prompt}
    \end{table*}

\section{Instructions}
\label{app:prompts}
We list all instructions used during training in Table \ref{tab:prompt}. \textsc{Argotario}, \textsc{Logic} and \textsc{Covid-19} have two instructions per example: \textit{List}: fallacy types listed in prompt, and \textit{Def} fallacy definitions included in prompt. For \textsc{Propaganda}, since each instance is sentence with a marked fallacious fragment, we construct three \textit{List} instructions where the fragment is included in the first instruction, removed completely in the second instruction (no fragment in Table \ref{tab:prompt}), and moved to the generation target in the third instruction (\textit{Frag}). The same three instructions are done using \textit{Def} prompts making the total six instructions per training example. For \textsc{Climate}, each instance is constructed using four instructions: \textit{List} and \textit{Def} with and without fact-checkers comments (\textit{Com}). These additional instructions for \textsc{Propaganda} and \textsc{Climate} are included during training only to increase the diversity of prompts.

Also, as discussed in \ref{sec:res}, we use few shot instruction-tuning of GPT-3 with and without explanations. The instructions that do not include explanations follow the same format of the \textit{List} prompts shown in Table \ref{tab:prompt} where it starts ``Given a text segment ...'' followed by a list of fallacy types and then the few shot examples that include a text segment and a fallacy type. Additionally, we write explanations after each few-shot example in the instruction prompt which explains why a given text segment is labeled with the fallacy type. The explanations follow the fallacy type labels as shown in Table \ref{tab:explanation}.

\end{document}